\documentclass[10pt,twocolumn,letterpaper]{article}

\usepackage[pagenumbers]{cvpr} 

\usepackage[dvipsnames]{xcolor}

\usepackage{graphicx, caption, amsmath, enumerate}
\usepackage{amssymb, mathtools, multirow, multicol}
\usepackage[colorinlistoftodos]{todonotes}
\usepackage{booktabs,colortbl}
\usepackage{gensymb}

\usepackage[accsupp]{axessibility} 

\definecolor{cvprblue}{rgb}{0.21,0.49,0.74}
\usepackage[pagebackref,breaklinks,colorlinks,citecolor=cvprblue]{hyperref}

\def\paperID{19} 
\def\confName{CVPR}
\def\confYear{2024}

\title{Robust and Explainable Fine-Grained Visual Classification with Transfer Learning: A Dual-Carriageway Framework}

\author{Zheming Zuo$^{1}$,\hspace{0.5em} Joseph Smith$^{1}$,\hspace{0.5em} Jonathan Stonehouse$^{2}$,\hspace{0.5em} Boguslaw Obara$^{1}$\\
$^{1}$School of Computing, Newcastle University, UK \hspace{1em} $^{2}$Procter and Gamble, UK\\
{\tt\small \{zheming.zuo,j.smith57,boguslaw.obara\}@newcastle.ac.uk \hspace{1em} stonehouse.jr@pg.com}
}

\begin{document}
\maketitle 
\begin{abstract}

In the realm of practical fine-grained visual classification applications rooted in deep learning, a common scenario involves training a model using a pre-existing dataset. Subsequently, a new dataset becomes available, prompting the desire to make a pivotal decision for achieving enhanced and leveraged inference performance on both sides: Should one opt to train datasets from scratch or fine-tune the model trained on the initial dataset using the newly released dataset? The existing literature reveals a lack of methods to systematically determine the optimal training strategy, necessitating explainability. To this end, we present an automatic best-suit training solution searching framework, the Dual-Carriageway Framework (DCF), to fill this gap. DCF benefits from the design of a dual-direction search (starting from the pre-existing or the newly released dataset) where five different training settings are enforced. In addition, DCF is not only capable of figuring out the optimal training strategy with the capability of avoiding overfitting but also yields built-in quantitative and visual explanations derived from the actual input and weights of the trained model. We validated DCF's effectiveness through experiments with three convolutional neural networks (ResNet18, ResNet34 and Inception-v3) on two temporally continued commercial product datasets. Results showed fine-tuning pathways outperformed training-from-scratch ones by up to 2.13\% and 1.23\% on the pre-existing and new datasets, respectively, in terms of mean accuracy. Furthermore, DCF identified reflection padding as the superior padding method, enhancing testing accuracy by 3.72\% on average. This framework stands out for its potential to guide the development of robust and explainable AI solutions in fine-grained visual classification tasks.
\end{abstract}

\section{Introduction}
\label{sec:intro}

Fine-Grained Visual Classification (FGVC) \cite{liu2021cross}, as a specialised task within the realm of computer vision, focuses on analysing the visual objects from subordinate categories with applications in biological research \cite{liu2023transifc}, automotive industry \cite{du2023multi}, fashion \cite{zhu2023learning} and retail \cite{sakai2023framework, min2023large}, etc. The past two decades have witnessed a decent amount of FGVC research outcomes \cite{lin2015bilinear, fu2017look, wang2018learning, du2020fine, du2021progressive, chang2023making} suppressing human experts since the broad application of Convolutional Neural Networks (CNNs) \cite{li2021survey, 10078845}.

\begin{figure}[!htbp]
	\centering
	\includegraphics[width=.85\linewidth]{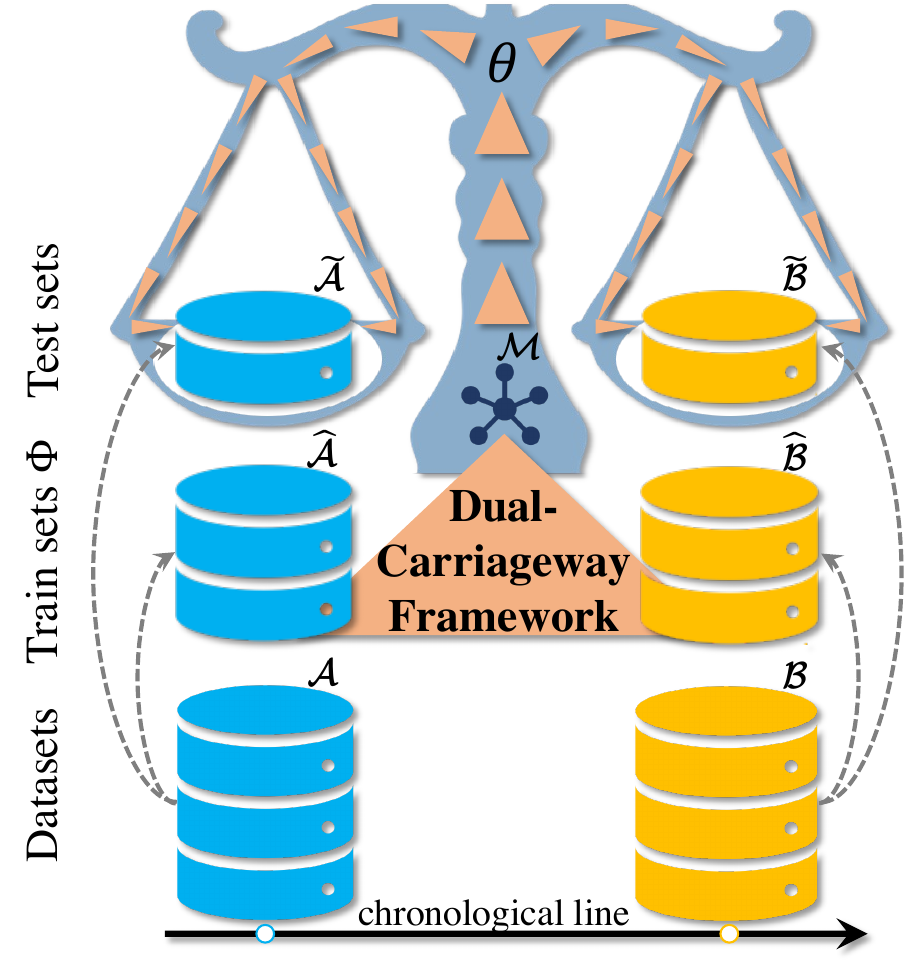}
	\caption{A robust and explainable learning framework should be revealed by not only the leveraged FGVC performance over two temporally continued datasets with the existence of subtle pattern differences, imbalanced data samples and high sparsity within the region of interest but also the associated quantitative (via the actual model input, \emph{i.e.}~frequency distribution of the padded image) and visual explanations (through the learned model weights, layer-wise attention) in a \emph{putting-through} manner.}
	\label{fig:mot}
\end{figure}

FGVC tends to be more challenging than conventional image classification due to the high degree of similar appearance among subordinate categories \cite{liu2022focus}. Besides, the coarse granularity of the captured image resulted from illumination variations (\emph{e.g.} low-light \cite{9049390} or over-exposed \cite{fu2023raw} environment) and camera motion associated with unstable hand stability (\emph{e.g.} blurriness \cite{zuo2022idea} and distortion \cite{li2023fg}) may deteriorate FGVC performance.

In a real-world scenario depicted in Fig.~\ref{fig:mot}, suppose that $\mathcal{A}$ and $\mathcal{B}$ are the two available datasets; this paper aims to find an optimal solution to leverage the prediction performance of a model $\mathcal{M}$ on their testing sets, \emph{i.e.} $\widetilde{\mathcal{A}}$ and $\widetilde{\mathcal{B}}$. Such an optimisation process can be expressed by:
\begin{equation}
\label{eq:opt}
\operatorname*{argmax}_{\theta} \left( \text{Acc}(\mathcal{M}_{\theta,\Phi}(\widetilde{\mathcal{A}})), \text{Acc}(\mathcal{M}_{\theta,\Phi}(\widetilde{\mathcal{B}}))
\right),
\end{equation}
where the optimal hyperparameters $\theta$ were learned by model $\mathcal{M}$ on the data $\Phi$. The above optimisation problem is usually resolved by training two datasets from scratch in a row \cite{wang2020pruning} or conducting transfer learning \cite{neyshabur2020being, zhuang2020comprehensive, niu2020decade}.

Fundamentally, the choice of training pathway highly relies on the data scale and pattern disparity between datasets $\mathcal{A}$ and $\mathcal{B}$. However, those characteristics cannot be precisely quantified as the existence of challenging factors mentioned before the antecedent paragraph and various types of uncertainties (\emph{e.g.}~dataset noise and model randomness) \cite{jungmann2024analytical} raised during the model training phase. The former would be less preferred if those two datasets had very different distributions. In contrast, the latter one suffers from the issue of catastrophic forgetting~\cite{boschini2022transfer,srivastava2023lifelong} where the model forgets important features from dataset $\mathcal{A}$ when fine-tuning on dataset $\mathcal{B}$. Besides, though transfer learning is a more common choice due to its much lower computational cost and possibly better generalisation capability contributed by \emph{e.g.} the large-scale ImageNet dataset \cite{5206848}, it may distort the pre-trained features \cite{kumar2022fine}.

The challenges mentioned above in selecting the appropriate training pathway for levering the prediction performance on $\widetilde{\mathcal{A}}$ and $\widetilde{\mathcal{B}}$ that are adjacent in the timeline as depicted in Fig.~\ref{fig:mot}, this work addresses the problem from a different yet explainable perspective. Generally, we present an automatic framework for selecting the best model training scheme, named Dual-Carriageway Framework (DCF), for an FGVC task given two chronologically continued datasets, and our contributions are summarised as follows:

\textbf{1)} We propose an automatic best-suit training solution searching framework (Sec.~\ref{sec:met}) through five training settings (Sec.~\ref{sec:model_train_settings}) configured in a dual-direction manner for robust fine-grained visual classification where the prediction performance on test sets of two temporally continued datasets are maximised.

\textbf{2)} We prove the feasibility and efficacy of the proposed framework (Sec.~\ref{sec: tans_solutions}) through not only the prediction accuracy and degree of confidence yielded by the three trained CNN models but also quantitative and visual explanations in a \emph{putting-through} manner from the actual model input data and layer-wise attention.

\textbf{3)} We show that determining the potentially most appropriate padding scheme (Sec.~\ref{sec:pad_schemes}) is a reasonable starting point (Sec.~\ref{sec:opt_pad_scheme_select}) for deploying the proposed framework, where merely the CNN model trained on the dataset collected earlier in the timeline is needed.


\section{Materials}
\label{sec:mat}

This section introduces the two datasets employed in this work, following the fine-grained classification task process sequence. Then, we combine several representative works to explain our choice and rationale for semantic segmentation models, padding schemes, and visual classifiers.

\subsection{Datasets}
\label{sec:ds}

\begin{figure}[!htbp]
	\centering
	\includegraphics[width=.98\linewidth]{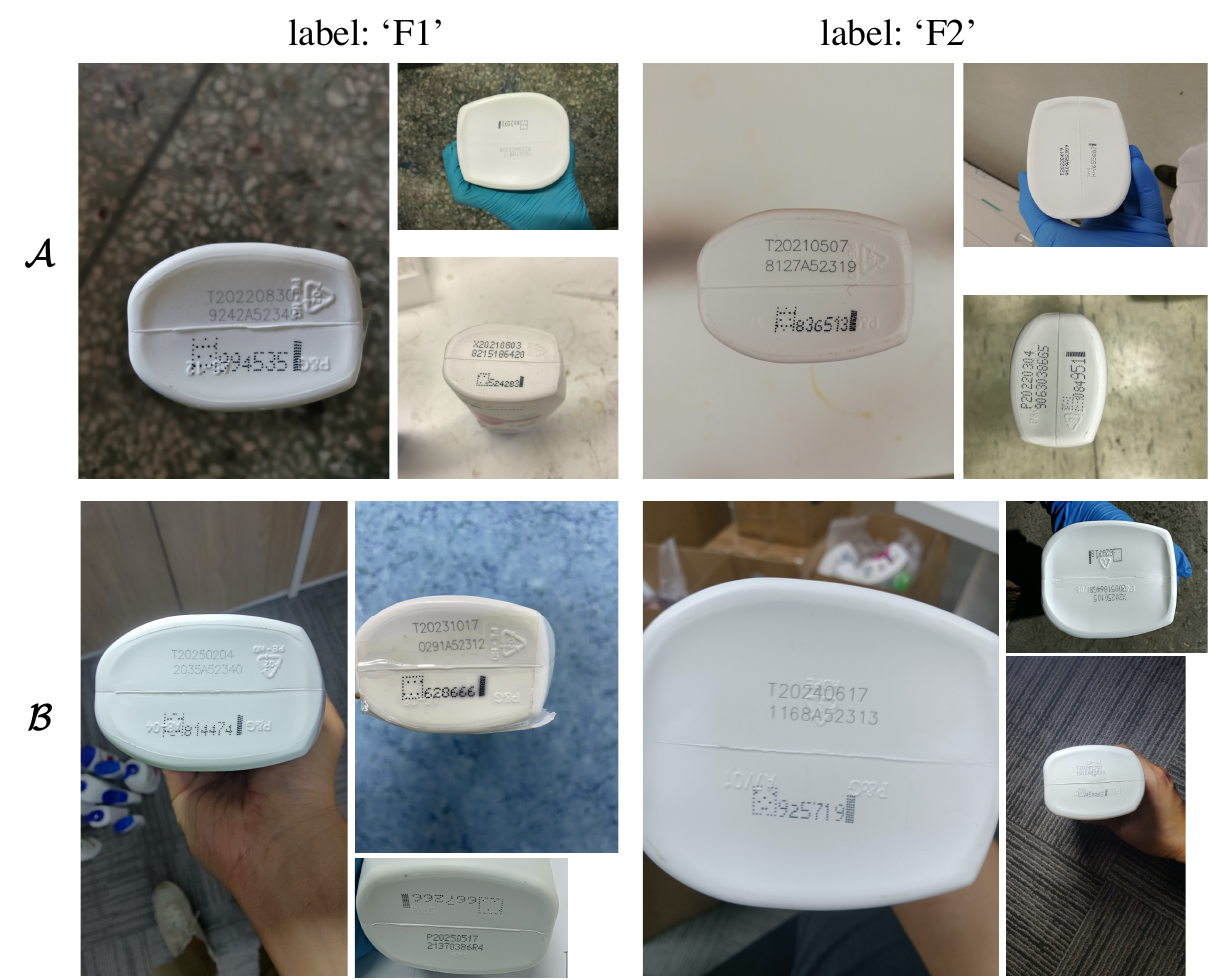}
	\caption{Sample images of the bottom side of commercial products produced by two manufacturers, `F1' and `F2', in pre-existing ($\mathcal{A}$) and newly released ($\mathcal{B}$) datasets. Target regions (dotted parts to be segmented by U-Net \cite{ronneberger2015u} in Sec.~\ref{sec:classify}) with ever-evolving anti-counterfeit code patterns embody spatial variations led by varying illumination, camera bias \cite{9534097}, and noisy textures \cite{jia2019coarse} (\emph{e.g.} dust and stains), affecting the overall FGVC accuracy.}
	\label{fig:data_sample}
\end{figure}

To facilitate the conduct of this research, our industrial partners delivered two RGB image datasets, $\mathcal{A}$ and $\mathcal{B}$, of the bottom side of commercial products, partially presented in Fig.~\ref{fig:data_sample} in chronological order. Each dataset contained two categories of instances indicating the corresponding manufacturers, labelled `F1' and `F2'.

Datasets $\mathcal{A}$ and $\mathcal{B}$ contain 5870 and 3112 RGB images, each of which comes with different resolutions, as the data-capturing process was performed by varying types of cameras in uncontrolled scenes. Slightly different from the commonly adopted data division approach, we use the data of the tail time block in both datasets as the test set, \emph{i.e.} the test set of $\mathcal{A}$ ($\widetilde{\mathcal{A}}$) contains 1716 images in which $|\widetilde{\mathcal{A}}_{\text{F1}}| = 1389$ and $|\widetilde{\mathcal{A}}_{\text{F2}}| = 327$, and the test set of $\mathcal{B}$ ($\widetilde{\mathcal{B}}$) includes 822 images where $|\widetilde{\mathcal{B}}_{\text{F1}}| = 220$ and $|\widetilde{\mathcal{B}}_{\text{F2}}| = 602$. The remaining part is divided into a training set and a validation set in a ratio of 4:1.

\subsection{Semantic Segmentation Models}
\label{sec:classify}

A substantial body of literature exists on the employment of deep neural networks for image semantic segmentation. Given that semantic segmentation is not the primary focus of this work, we briefly review key network architectures in the order of their introduction and describe our choice.

Fully Convolutional Network (FCN) \cite{long2015fully}, as an early seminal contribution to semantic segmentation, is a simple yet efficient architecture that preserves spatial information through end-to-end pixel-wise predictions due to the substitution of fully connected layers using the convolutional ones. As the evolutionary successor to FCN, U-Net \cite{ronneberger2015u} demonstrates a sophisticated capacity for capturing coarse and fine-grained features, achieving a favourable balance between temporal efficiency and richer feature representation. Its distinctive U-shaped architectural configuration integrates contracting and expansive pathways \cite{fu2022x,xu2022omega}, culminating in enhanced accuracy in spatial localisation.

DeepLab~\cite{chen2017deeplab} utilises atrous convolutions and spatial pyramid pooling to capture multi-scale contextual information, thus enhancing the model's ability to consider a broader context and overcome the restricted receptive field limitation associated with U-Net. High-Resolution Network (HRNet)~\cite{wang2020deep} mitigates DeepLab's limitations by devising high-resolution feature fusion, enhancing accuracy in fine details without compromising computational efficiency. Such a fusion addresses challenges related to longer training times and limited generalisation observed in DeepLab.

As the region of interest in this work is the anti-counterfeit code (with significant difference relative to background) in each of the given raw RGB images (depicted in the dotted part of Fig.~\ref{fig:data_sample}), a U-Net \cite{ronneberger2015u}, which is pre-trained on 1479 images with bounding boxes that annotated by the Amazon Mechanical Turk (MTurk) \cite{mturk} crowdsourcing platform and refined by the Maximally Stable Global Region \cite{jia2019coarse}, is employed in this work for semantic segmentation (\emph{i.e.} the input of the padding schemes partially introduced in Sec.~\ref{sec:pad_sch} and systematically formulated in Sec.~\ref{sec:pad_schemes}), as exampled in the leftmost side of Fig.~\ref{fig:note}, to facilitate the nature of an FGVC task.

\subsection{Padding Schemes and Classification Models}
\label{sec:pad_sch}

There is a limited amount of literature that proposes or systematically compares \cite{amam2021position} padding schemes in FGVC, except for the most commonly adopted \texttt{zero} padding \cite{norouzi2009stacks} in the past decade. Noteworthy, while many works focus on how to make neural networks deeper or wider, the padding scheme, which is usually neglected, is also closely related to the FGVC performance \cite{lopez2020effect}. In CNNs \cite{726791}, \texttt{zero} padding maintains the spatial size of the output feature maps by adding layers of zeros around the input image before performing the convolution operation. However, it may introduce artefacts into the data that the model learns, leading to possible overfitting or reduced model generalisability \cite{ke2023s}. The \texttt{reflection} padding addresses this by mirroring the image at the borders, maintaining the continuity of image features and reducing the edge artefacts \cite{engstrom2019rotation,chen2023residual}. However, it may introduce redundancy and may not be suitable for all types of images. Similarly, padding schemes based on partial convolution \cite{liu2018partialpadding,liu2018image} and repetition \cite{nkwentsha2020automatic} are robust to boundary artefacts \cite{liu2022partial}.

On the contrary, the learning-based \cite{ke2023s,alrasheedi2023padding}, interpolation-based \cite{hashemi2019enlarging}, and distribution-based \cite{nguyen2019distribution} padding schemes focus on leveraging the connectivity of image borders for more effective padding to reflect the natural extension of the image content and suppress the risk of distorting the convolution. In addition, a padding scheme dynamically adjusted based on the contextual information around the border pixels is devised explicitly for semantic segmentation \cite{he2021cap}.

Furthermore, padding with a non-zero constant value is also a viable option, albeit infrequently implemented in practical applications \cite{nguyen2019distribution}. Exploring whether segmented rectangular images and padded parts are complementary or mutually exclusive regarding model perceptibility would be interesting. To this end, considering that the segmented images are relatively sparse, and their width is much greater than their height, we systematically formulate six padding schemes, including \texttt{zero} padding and three schemes of padding with a non-zero constant in Eq.~(\ref{eq:non_zero_cons_pad}) as well as the \texttt{reflection} padding Eq.~(\ref{eq:ref_padding}) in Sec.~\ref{sec:pad_schemes}.

The padded images with the resolution of 1024-by-1024 pixels, as the actual input to CNNs, are then readily fed forward for convolution operations. To perform the binary classification task, we adopt three CNNs for binary classification, \emph{i.e.}~two residual networks (ResNet18 and ResNet34) \cite{he2015deep} and one Inception network (Inception-v3) \cite{szegedy2016rethinking} due to their advantages of overcoming vanishing gradients \cite{oyedotun2022everyone} and capturing multi-scale features \cite{zhang2022multi}. To take advantage of large-scale model pretraining, all three models initialise the training process based on the weights derived from ImageNet \cite{5206848}. Detailed training pathways are elaborated in Sec.~\ref{sec:model_train_settings}, and setups are configured in Sec.~\ref{sec:setups}.

\section{The Dual-Carriageway Framework}
\label{sec:met}

In this section, we present the proposed DCF for FGVC in detail. First, we briefly explain the DCF workflow and introduce its two key components.

\subsection{Workflow}

\begin{figure*}[ht]
  \centering
  \includegraphics[width=0.98\linewidth]{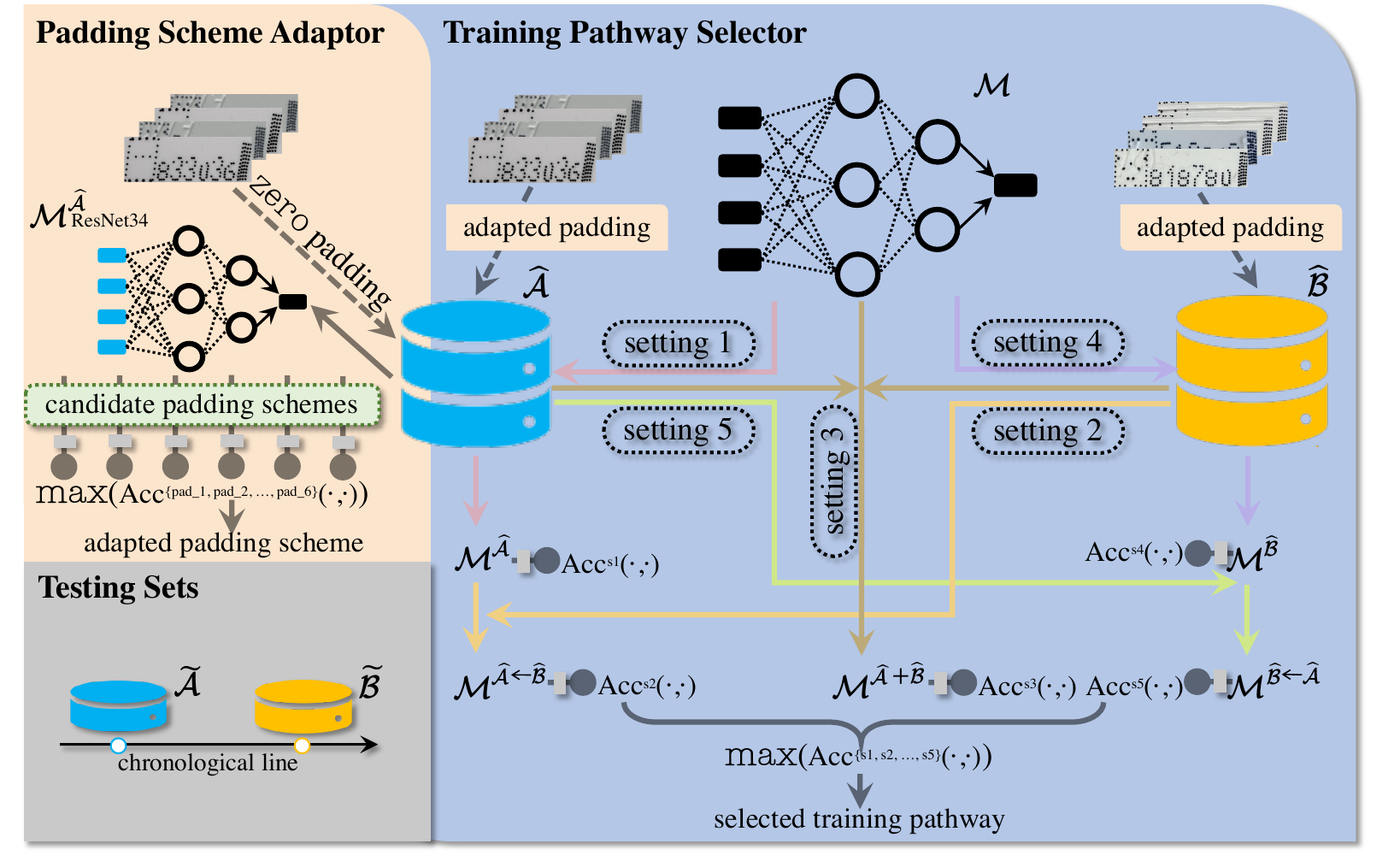}
  \caption{Workflow of the proposed DCF for robust and explainable fine-grained visual classification. This figure is separated into two consecutive components and a shared part by colours corresponding to 1) the Padding Scheme Adaptor (Sec.~\ref{sec:pad_schemes}), 2) the Training Pathway Selector (Sec.~\ref{sec:model_train_settings}), and 3) testing sets with pattern variations derived from two chronologically continued datasets (Sec.~\ref{sec:ds}).}
   \label{fig:note}
\end{figure*}

The workflow of the proposed DCF is depicted
in Fig. \ref{fig:note}. Briefly, it contains two consecutive components: a Padding

\noindent Scheme Adaptor (PSA) (detailed in the upper left side of Fig.~\ref{fig:note}) and a Training Pathway Selector (TPS) (depicted in the right-hand side of Fig.~\ref{fig:note}). Concretely, PSA is proposed to figure out the appropriate padding scheme to be adopted in the subsequent TPS where the FGVC accuracies on both temporally continued datasets (visualised in the lower left side of Fig.~\ref{fig:note}) are maximised and leveraged.

In PSA, we first train a ResNet34 \cite{he2015deep} model (one of the three visual classification models adopted in this work and mentioned in the tail part of Sec.~\ref{sec:pad_sch}) on the earlier training set $\widehat{\mathcal{A}}$ (splitter from its full dataset, $\mathcal{A}$ detailed in Sec.~\ref{sec:ds}) with the most commonly employed \texttt{zero} padding (revisited in the first half of Sec.~\ref{sec:pad_sch}). Such a trained model is tested on each of the six padding schemes (visualised in Fig.~\ref{fig:ava_pad}) that are systematically formulated in Sec.~\ref{sec:pad_schemes}, using Eqs.~(\ref{eq:non_zero_cons_pad}) and (\ref{eq:ref_padding}) in general and Eqs.~(\ref{eq:non_zero_cons_pad_value}) and (\ref{eq:ref_padding_value}) in specific. Given that subtle changes in terms of the anti-counterfeit code pattern existed in datasets $\mathcal{A}$ and $\mathcal{B}$, we can retain the most suitable padding scheme in our latter TPS phase by looking for the maximised inference performance on both testing sets. In TPS, for each of three visual classifiers (ResNet18, ResNet34 and Inception-v3 \cite{szegedy2016rethinking} adopted, we perform model training using five settings that are introduced in Sec.~\ref{sec:model_train_settings} and summarised in Eq.~(\ref{eq:setups}) in a dual-carriageway manner, \emph{i.e.} train on $\widehat{\mathcal{A}}$ only, fine-tune the model learned from $\widehat{\mathcal{A}}$ using $\widehat{\mathcal{B}}$, train on both $\widehat{\mathcal{A}}$ and $\widehat{\mathcal{B}}$, train on $\widehat{\mathcal{B}}$ only, and fine-tune the model learned from $\widehat{\mathcal{B}}$ using $\widehat{\mathcal{A}}$.

\subsection{Padding Scheme Adaptor}
\label{sec:pad_schemes}

\begin{figure}[!ht]
  \centering
  \includegraphics[width=0.98\linewidth]{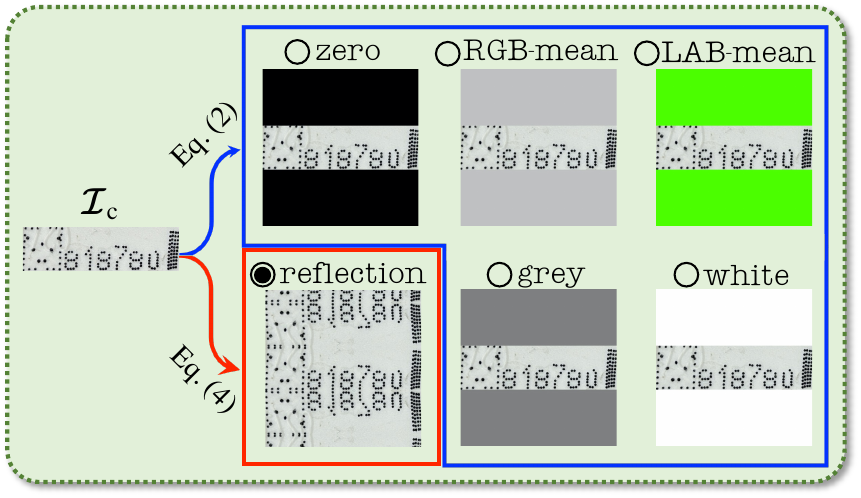}
  \caption{Candidate padding schemes available in the PSA component of the proposed DCF presented in Fig.~\ref{fig:note}.}
   \label{fig:ava_pad}
\end{figure}

As the output of a pre-trained semantic segmentation model \cite{ronneberger2015u, jia2019coarse} mentioned in Sec.~\ref{sec:classify}, each cropped image is $\mathcal{I}_{\text{c}}$ in the shape of a rectangle whose width is much larger than its height. Given the characteristics of a range of padding schemes that have been systematically introduced in Sec.~\ref{sec:pad_sch}, we determine that a total of six padding schemes (in Fig.~\ref{fig:ava_pad}), \emph{i.e.}~three constant value-based ones (\texttt{zero}, \texttt{white} \cite{norouzi2009stacks} and \texttt{grey}), two mean value-based ones (\texttt{RGB-mean} and \texttt{LAB-mean}), and one mirroring-based method (\texttt{reflection}) (a variant of \cite{engstrom2019rotation} and \cite{chen2023residual}), are evenly performed in the vertical direction merely (except the \texttt{reflection} padding) with three practical advantages: a) preserving different aspect ratios; b) avoiding unnecessary distortions and c) reducing the computational cost.

Concretely, given a cropped image $\mathcal{I}_{\text{c}}\in \mathbb{R}^{H \times W \times C}$ where $H\ll W$, the \texttt{zero}, \texttt{RGB-mean}, \texttt{LAB-mean}, \texttt{white} and \texttt{grey} padding schemes are performed evenly on both top and bottom side of $\mathcal{I}_{\text{c}}$, \emph{i.e.}~$P_{\text{t}} = P_{\text{b}}$, and thus the padded image $\mathcal{I}_{\text{p}} \in \mathbb{R}^{W \times W \times C}$ is defined by:

\begin{equation}
\label{eq:non_zero_cons_pad}
    \mathcal{I}_{\text{p}}[i^\prime, j, :] =
\begin{cases}
    \mathcal{I}_{\text{c}}[i-P_{\text{t}}, j, :] & \text{if } i \in [P_{\text{t}}, H + P_{\text{t}}),  \\
    (v_1, v_2, v_3) & \text{otherwise},
\end{cases}
\end{equation}
where $i \in [0, H-1]$, $j \in [0, W-1]$, $i^\prime \in [0, H+P_{\text{t}}-1]$ s.t. $P_{\text{t}} = P_{\text{b}} = (W-H)/2$, and $(v_1, v_2, v_3)$ is computed as:

\begin{equation}
\label{eq:non_zero_cons_pad_value}
\begin{cases}
    (0, 0, 0) & \texttt{zero}, \\
    \frac{1}{HW} \sum_{i=0}^{H-1} \sum_{j=0}^{W-1} \mathcal{I}_{\text{c}}[i, j, :] & \texttt{RGB-mean}, \\
    \frac{1}{HW} \sum_{i=0}^{H-1} \sum_{j=0}^{W-1} \mathcal{I}_{\text{c} \rightarrow \text{LAB}}[i, j, :] & \texttt{LAB-mean}, \\
    (255, 255, 255) & \texttt{white}, \\
    (128, 128, 128) & \texttt{grey}. \\
\end{cases}
\end{equation}

Particularly, the \texttt{reflection} padding scheme is expressed as:

\begin{equation}
\label{eq:ref_padding}
    \mathcal{I}_{\text{p}}[i^\prime, j, :] = \mathcal{I}_{\text{c}}[\texttt{reflect}(I, P_{\text{t}}, H), j, :],
\end{equation}
in which the function $\texttt{reflect}(\cdot)$ is conducted when the index $i$ meets in the following criteria in the vertical direction of $\mathcal{I}_{\text{c}}$:

\begin{equation}
\label{eq:ref_padding_value}
\resizebox{.905\linewidth}{!}{%
    $\texttt{reflect}(i, P_{\text{t}}, H) =
    \begin{cases}
        i & \text{if } i \in [0, H), \\
        2P_{\text{t}}-i-1 & \text{if } i \in [H, H+P_{\text{t}}].
    \end{cases}$
}
\end{equation}

This ensures that \texttt{reflection} padding occurs when the index $i$ is beyond the top border of $\mathcal{I}_{\text{c}}$. Additionally, if $H < P_{\text{t}}$, multiple times of refections will occur, simulating the effect of extending the height of $\mathcal{I}_{\text{c}}$.

\subsection{Training Pathway Selector}
\label{sec:model_train_settings}

To resolve the optimisation problem defined in Eq.~(\ref{eq:opt}), given the training sets $\widehat{\mathcal{A}}$ and $\widehat{\mathcal{B}}$ split by the ratio defined in Sec.~\ref{sec:ds}, optimal hyperparameters $\theta$ are commonly learned through the following dual-carriageway forms:

\begin{equation}\label{eq:setups}
\theta \vcentcolon= \begin{cases}
\mathcal{M}^{\widehat{\mathcal{A}}} & \text{setting}\,\, 1,\\
\mathcal{M}^{\widehat{\mathcal{A}} \leftarrow \widehat{\mathcal{B}}} & \text{setting}\,\, 2,\\
\mathcal{M}^{\widehat{\mathcal{A}} + \widehat{\mathcal{B}}} & \text{setting}\,\, 3,\\
\mathcal{M}^{\widehat{\mathcal{B}}} & \text{setting}\,\, 4,\\
\mathcal{M}^{\widehat{\mathcal{B}} \leftarrow \widehat{\mathcal{A}}} & \text{setting}\,\, 5,\\
\end{cases}
\end{equation}
in which $\mathcal{M}^{\widehat{\mathcal{A}} + \widehat{\mathcal{B}}}$ denotes the model trained on $\widehat{\mathcal{A}}$ and $\widehat{\mathcal{B}}$ together, whereas $\mathcal{M}^{\widehat{\mathcal{A}} \leftarrow \widehat{\mathcal{B}}}$ represents the model trained on $\widehat{\mathcal{A}}$ first and then fine-tuned using $\widehat{\mathcal{B}}$, and vice versa. In addition, the alias of each training setting is used in the bottom right side of Fig.~~\ref{fig:note}, \emph{e.g.}~s1 denotes setting 1.

Specifically, we adopt three relatively light CNN models in this work to justify the proposed framework of seeking an optimal training pathway for robust and explainable FGVC. Noteworthy, for training settings 2 and 4 of Eq.~(\ref{eq:setups}), we fine-tune each of the CNNs by layer freezing for simplicity and time-efficiency \cite{shen2021partial,shi2023towards,sarfi2023simulated}. Detailed experimental setups are presented in Sec.~\ref{sec:setups}.

\section{Experiments}
\label{sec:exp}

\subsection{Experimental Setups and Evaluation Metrics}
\label{sec:setups}

We implemented DCF on an NVIDIA RTX 3090 Ti GPU. To ensure uniform input image size of deep neural classifiers and compensate for various aspect ratios of the segmented images, we scale them to a resolution of $1024\times 1024$ pixels with six padding schemes (detailed in Sec.~\ref{sec:pad_schemes}) applied. For training, each of the three image classifiers (Sec.~\ref{sec:classify}), we adopt Adam optimiser \cite{kingma2014adam} and categorical cross-entropy loss \cite{sukhbaatar2014learning} with a learning rate of 0.0001 and a weight decay of 0. Besides, we adopt random rotation for data augmentation to improve model generalisation.

The result was reported as mean accuracy $\pm$ standard deviation by training the same deep image classifier five times for a fair comparison of FGVC performance. As a valuable add-on to the proposed DCF, we provide quantitative and visual explanations of the padded input (\emph{i.e.} frequency distribution) and weights (\emph{i.e.} layer-wise attention) associated with the trained classifier via mean pixel value and Gradient-weighted Class Activation Mapping (GradCAM) \cite{selvaraju2017grad}. More experimental results are available in our \emph{supplementary materials}.

\subsection{Optimal Padding Scheme Determination}
\label{sec:opt_pad_scheme_select}

As briefly introduced (Sec.~\ref{sec:pad_sch}) and systematically designed (Sec.~\ref{sec:pad_schemes}), we trained a ResNet34 model with the most com-

\begin{figure*}[!htbp]
    \centering
    \begin{minipage}{0.5\textwidth}
        \centering
        \captionof{table}{Prediction performance (in \%) comparisons on the testing sets of datasets $\mathcal{A}$ (\emph{i.e.} $\widetilde{\mathcal{A}}$) and $\mathcal{B}$ (\emph{i.e.} $\widetilde{\mathcal{B}}$) using the baseline models (ResNet34) trained on the training set of dataset $\mathcal{A}$ (\emph{i.e.} $\widehat{\mathcal{A}}$) that was processed by various padding schemes. Results yielded by the optimal padding scheme are marked in bold and highlighted with \colorbox{gray!30}{\phantom{zz}} (detailed in Table \ref{tab:reflec_pad_details}).}
        \label{tab:opti_pad_selec}
        \scalebox{0.85}{
        \begin{tabular}{lcc}
        \toprule
        Padding Scheme & Acc\big($\mathcal{M}_{\text{ResNet34}}^{\widehat{\mathcal{A}}}(\widetilde{\mathcal{A}})$\big) & Acc\big($\mathcal{M}_{\text{ResNet34}}^{\widehat{\mathcal{A}}}(\widetilde{\mathcal{B}})$\big)\\
        \midrule
        \texttt{zero} & 94.05 $\pm$ 2.03 & 71.79 $\pm$ 3.15\\
        \texttt{RGB-mean} & 95.07 $\pm$ 0.53 & 74.59 $\pm$ 0.86\\
        \texttt{LAB-mean} & 95.59 $\pm$ 0.48 & 72.61 $\pm$ 1.21\\
        \texttt{white} & 94.89 $\pm$ 0.20 & 75.78 $\pm$ 1.41\\
        \texttt{grey} & 95.10 $\pm$ 0.41 & 77.10 $\pm$ 1.06\\
        \texttt{reflection} & \cellcolor{gray!30}\textbf{96.58 $\pm$ 0.35} & \cellcolor{gray!30}\textbf{79.85 $\pm$ 1.18}\\
        \bottomrule
    \end{tabular}}
        
        \vspace{1mm} 
        
        \captionof{table}{Detailed prediction performance (in \%) of ResNet34 trained on $\widehat{\mathcal{A}}$ with \texttt{reflection} padding. The optimal model is marked in bold and highlighted by \colorbox{gray!30}{\phantom{zz}} (adopted for tuning purposes in Sec.~\ref{sec: tans_solutions}).}
        \label{tab:reflec_pad_details}
        \scalebox{0.85}{
    \begin{tabular}{lcc}
        \toprule
        run & Acc\big($\mathcal{M}_{\text{ResNet34}}^{\widehat{\mathcal{A}}}(\widetilde{\mathcal{A}})$\big) & Acc\big($\mathcal{M}_{\text{ResNet34}}^{\widehat{\mathcal{A}}}(\widetilde{\mathcal{B}})$\big)\\
        \midrule
        1 & 97.09 & 79.81\\
        2 & 96.68 & 78.98\\
        \cellcolor{gray!30}\textbf{3} & \cellcolor{gray!30}\textbf{96.60} & \cellcolor{gray!30}\textbf{81.87}\\
        4 & 96.19 & 79.04\\
        5 & 96.32 & 79.56\\
        \bottomrule
    \end{tabular}}
    \end{minipage}%
    \hspace{1mm}
    \begin{minipage}{0.42\textwidth}
        \centering
        \includegraphics[width=\linewidth]{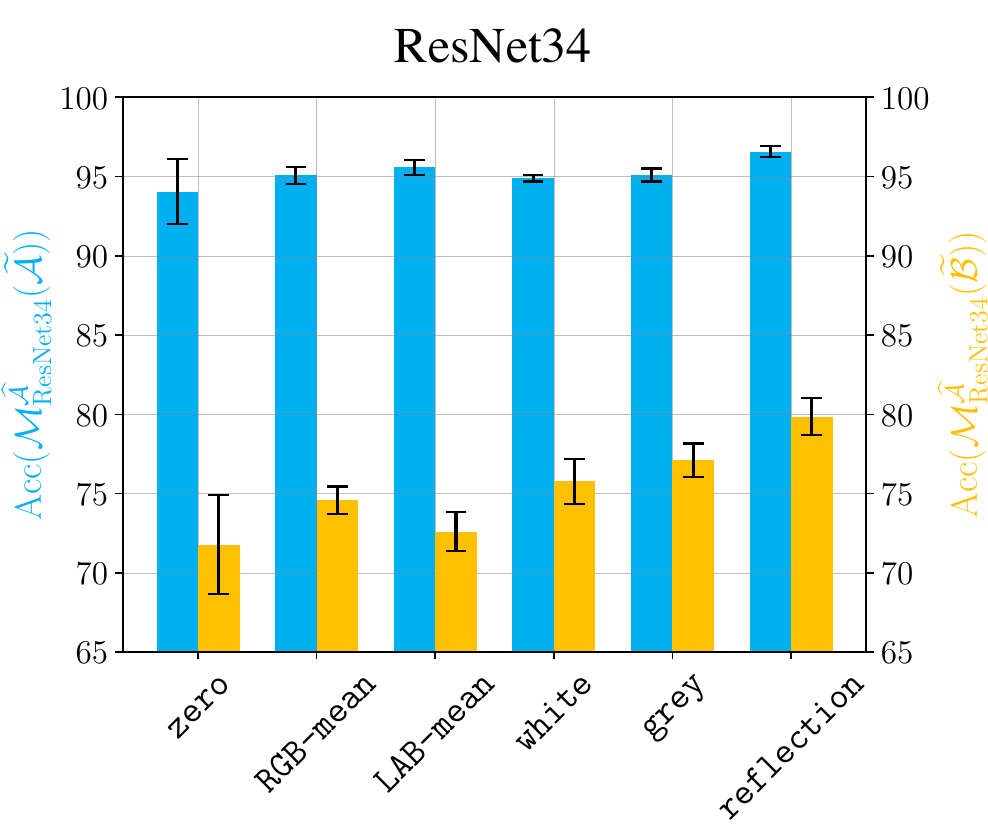}
        \captionof{figure}{Prediction performance (in \%) comparisons on the testing sets of datasets $\mathcal{A}$ (in the $y$-axis on the left) and $\mathcal{B}$ (in the $y$-axis on the right) using the baseline models (ResNet34) trained on the training set of dataset $\mathcal{A}$ that was processed by six padding schemes summarised in Table~\ref{tab:opti_pad_selec}. Results yielded by the optimal padding scheme -- \texttt{reflection} -- are detailed in Table \ref{tab:reflec_pad_details}. Quantitative and visual explanations of the six padding schemes are shown in Fig.~\ref{fig:padding_exp}. Colour codes are consistent with Figs.~\ref{fig:mot} and~\ref{fig:note}.}
        \label{fig:opti_pad_selec}
    \end{minipage}

    \vspace{1mm}
    \begin{minipage}{0.98\textwidth}
        \centering
        \captionof{table}{Quantitative explanation of the relationship between input pixel value and confidence of correct predictions given by ResNet34 trained with \texttt{zero} padding under six padding schemes. $avg(\cdot; \cdot)$ calculates the average for each of the two elements regarding the correctly predicted samples of `F1' and `F2' within the testing set, $\bar{p}$ (normalised to the range of $[0,1]$) represents the mean pixel value, $\mathcal{M}_{c}$ denotes the model confidence and $\mathcal{M}_{\text{acc}}$ corresponds to the model inference accuracy measured in \%. The best adapted $avg(\bar{p})$ values and their associated $\mathcal{M}_{c}$ ones are marked in bold and highlighted with \colorbox{gray!30}{\phantom{zz}}. Further visual explanation is delivered in Fig.~\ref{fig:padding_exp}.}
        \label{tab:opti_pad_selec_quant}
        \aboverulesep=0.1ex
        \belowrulesep=0.15ex
        \scalebox{0.87}{
        \setlength{\tabcolsep}{2pt}
        \begin{tabular}{lc|cc|cc|cc|cc|cc|cc}
        \toprule
        \multirow{2}{*}{Testing Set} & \multirow{2}{*}{Label} & \multicolumn{2}{c|}{\texttt{zero}} & \multicolumn{2}{c|}{\texttt{RGB-mean}} & \multicolumn{2}{c|}{\texttt{LAB-mean}} & \multicolumn{2}{c|}{\texttt{white}} & \multicolumn{2}{c|}{\texttt{grey}} & \multicolumn{2}{c}{\texttt{reflection}}\\
        & & $avg(\bar{p}; \mathcal{M}_{c})$ & $\mathcal{M}_{\text{acc}}$ & $avg(\bar{p}; \mathcal{M}_{c})$ & $\mathcal{M}_{\text{acc}}$ & $avg(\bar{p}; \mathcal{M}_{c})$ & $\mathcal{M}_{\text{acc}}$ & $avg(\bar{p}; \mathcal{M}_{c})$ & $\mathcal{M}_{\text{acc}}$ & $avg(\bar{p}; \mathcal{M}_{c})$ & $\mathcal{M}_{\text{acc}}$ & $avg(\bar{p}; \mathcal{M}_{c})$ & $\mathcal{M}_{\text{acc}}$\\
        \midrule
        \multirow{2}{*}{$\widetilde{\mathcal{A}}$} & F1 & 0.18; 0.989 & \multirow{2}{*}{95.22} & 0.60; 0.979 & \multirow{2}{*}{95.37} & 0.39; 0.986 & \multirow{2}{*}{94.98} & 0.88; 0.990 & \multirow{2}{*}{95.07} & 0.53; 0.976 & \multirow{2}{*}{94.91} & \cellcolor{gray!30}\textbf{0.60}; \textbf{0.990} & \multirow{2}{*}{96.60}\\
        & F2 & 0.18; 0.976 & & 0.60; 0.967 & & 0.33; 0.984 & & 0.88; 0.973 & & 0.53; 0.972 & & \cellcolor{gray!30}\textbf{0.60}; \textbf{0.983} &\\
        \midrule
        \multirow{2}{*}{$\widetilde{\mathcal{B}}$} & F1 & 0.20; 0.958 & \multirow{2}{*}{73.68} & 0.70; 0.941 & \multirow{2}{*}{75.66} & 0.58; 0.951 & \multirow{2}{*}{74.66} & 0.91; 0.961 & \multirow{2}{*}{77.37} & 0.56; 0.959 & \multirow{2}{*}{78.22} & \cellcolor{gray!30}\textbf{0.70}; \textbf{0.961} & \multirow{2}{*}{81.87}\\
        & F2 & 0.23; 0.909 & & 0.71; 0.917 & & 0.60; 0.932 & & 0.90; 0.932 & & 0.57; 0.952 & & \cellcolor{gray!30}\textbf{0.70}; \textbf{0.940} &\\
        \bottomrule
    \end{tabular}
    }
    \end{minipage}
\end{figure*}

\noindent monly employed \texttt{zero} padding scheme on the training split of a pre-existing dataset $\widehat{\mathcal{A}}$ as the backbone model, \emph{i.e.}~training setting 1 defined in Eq.~({\ref{eq:setups}}) and visualised in Fig.~{\ref{fig:note}}. From there, we evaluate the transferrable capability of such a model by adopting six padding schemes (include \texttt{zero} padding) on the testing split of the pre-existing and lastly released dataset, \emph{i.e.}~$\widetilde{\mathcal{A}}$ and $\widetilde{\mathcal{B}}$.

Testing performance on these two tests is summarised in Table~\ref{tab:opti_pad_selec} and visualised in Fig.~\ref{fig:opti_pad_selec}. It could be generally observed from Table~\ref{tab:opti_pad_selec} that \texttt{reflection} padding contributed the most to the inference performance compared to the rest, which is consistent with the findings presented in~\cite{amam2021position}. When testing above-mentioned backbone model on $\widetilde{\mathcal{A}}$, \emph{i.e.}~the column of Acc($\mathcal{M}_{\text{ResNet34}}^{\widehat{\mathcal{A}}}(\widetilde{\mathcal{A}})$) in Table~\ref{tab:opti_pad_selec}, the \texttt{reflection} padding outperforms \texttt{zero}, \texttt{RGB-mean}, \texttt{LAB-mean}, \texttt{white} and \texttt{grey} padding schemes by 2.53\%, 1.51\%, 0.99\%, 1.69\% and 1.48\% in terms of mean accuracy. Impressively, the \texttt{reflection} padding exceeds its counterparts with a more significant margin, \emph{i.e.}~8.06\%, 5.26\%, 7.24\%, 4.07\% and 2.75\% improvements, respectively, for mean accuracy. 

Furthermore, for selecting a single model to continue the conduction of the proposed DCF, we decompose the highlighted performance yielded by \texttt{reflection} padding (available in Table~\ref{tab:opti_pad_selec}) in Table~\ref{tab:reflec_pad_details}. One step further, in line with the columns containing `$\mathcal{M}_{\text{acc}}$' values as summarised in Table~\ref{tab:opti_pad_selec_quant}, the peak inference performance of the backbone model on two testing sets resulting from \texttt{reflection} padding is 96.60\% and 81.87\%, as outperformed \texttt{zero} padding for 1.38\% and 8.19\%, \texttt{RGB-mean} padding for 1.23\% and 6.21\%, \texttt{LAB-mean} padding for 1.71\% and 7.21\%, \texttt{white} padding for 1.53\% and 4.5\%, and \texttt{grey} pa-


\begin{figure*}[!htbp]
    \centering

    \begin{minipage}{0.94\textwidth}
        \centering
        \includegraphics[width=.94\linewidth]{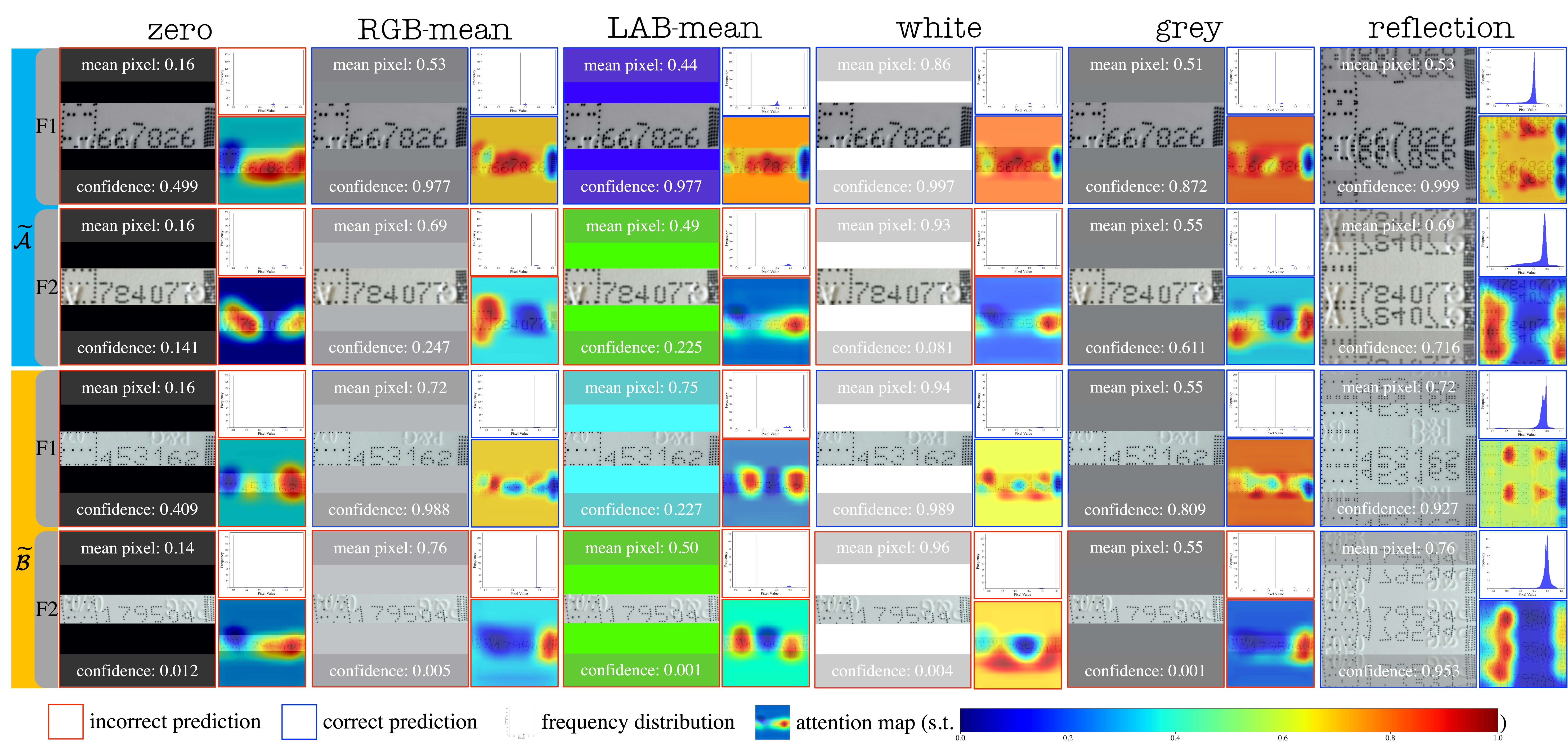}
        \captionof{figure}{The relationship between the input image with six different padding schemes applied and the model (ResNet34) predictions in the binary FGVC task is revealed by the frequency distribution and model prediction confidence in the testing sets of $\mathcal{A}$ and $\mathcal{B}$ (Sec.~\ref{sec:ds}), \emph{i.e.}~$\widetilde{\mathcal{A}}$ and $\widetilde{\mathcal{B}}$ with consistent colour schemes adopted in Figs.~\ref{fig:mot} and~\ref{fig:note}. Best viewed in colour and zoomed mode.}
        \label{fig:padding_exp}
    \end{minipage}

    \vspace{1mm}
    \begin{minipage}{0.94\textwidth}
        \centering
        \includegraphics[width=.94\linewidth]{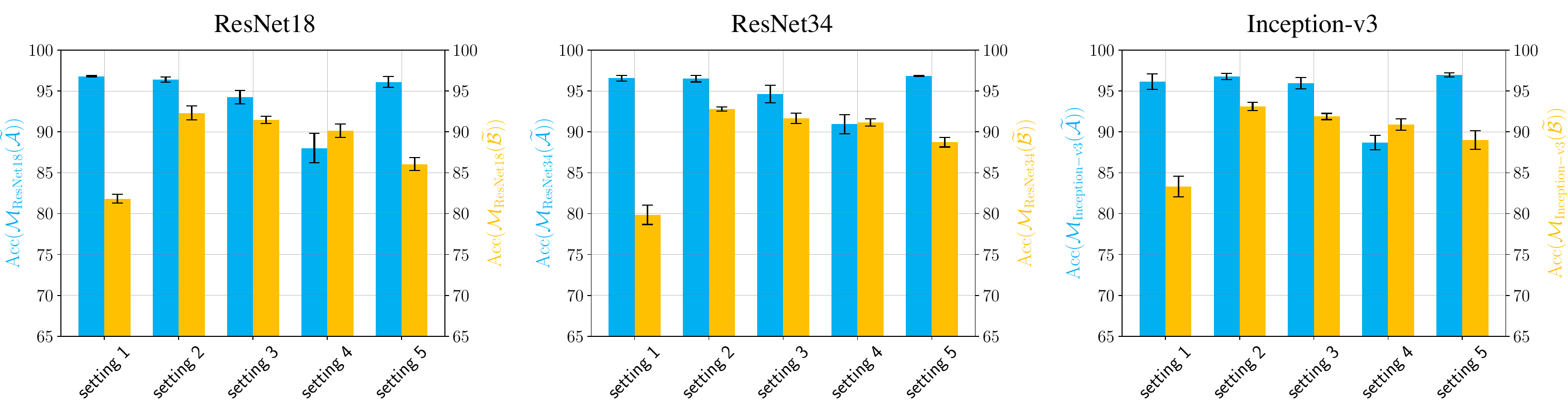}
        \captionof{figure}{Prediction performance (in \%) comparisons on the testing sets of datasets $\mathcal{A}$ (\emph{i.e.} $\widetilde{\mathcal{A}}$) and $\mathcal{B}$ (\emph{i.e.} $\widetilde{\mathcal{B}}$) using the three deep neural networks under five training settings with \texttt{reflection} padding scheme applied. This figure is a visual presentation of Table~\ref{tab:compare_transfer}. Colour codes are consistent with Figs.~\ref{fig:mot} and~\ref{fig:note}.}
        \label{fig:comparative_transfer}
    \end{minipage}
\end{figure*}
  
\noindent dding for 1.78\% and 3.65\%. Noteworthy, the above-peak performances are qualified following the criteria defined in Eq.~(\ref{eq:opt}), \emph{i.e.} the leveraged performance over two testing sets using a single model as opposed to ones that performed exceptionally well on one of the two testing sets and marginally poorly on another.

Besides, Table~\ref{tab:opti_pad_selec_quant} presented a quantitative explanation of the rationale that \texttt{reflection} padding outperforms its counterpart schemes. Concretely, the averaged mean pixel value of images padded using \texttt{reflection} padding in both subsets of `F1' and `F2' are constantly identical in the testing sets of $\widetilde{\mathcal{A}}$ and $\widetilde{\mathcal{B}}$, \emph{i.e.}~0.60 and 0.78, respectively. Such an observation confirms the robustness of the \texttt{reflection} padding schemes in the context of changing anti-counterfeit code patterns. We also provide visual explanations in Fig.~\ref{fig:padding_exp} to justify this. Overall, the competitive performance is mainly due to the backbone model's high degree of generalisation capability. This, in turn, reveals the practical feasibility and advantages of dense input regarding the more robust averaged mean pixel value, the corresponding more even frequency distribution, and the correctness of model attention.


\subsection{Optimal Training Pathway Selection} 
\label{sec: tans_solutions}

In this experiment phase, we assess the feasibility of identifying the most suitable training pathway selection from two sequential datasets, as delineated in Sec.~\ref{sec:model_train_settings}. Prediction performances on two datasets are summarised in Table~\ref{tab:compare_transfer} and visualised in Fig.~\ref{fig:comparative_transfer}.

\begin{table*}[ht]
\begin{minipage}[b]{0.35\linewidth}\centering
\caption{Summary of prediction performance yielded by three fine-grained visual classifiers under five training pathways that defined in Eq.~(\ref{eq:setups}). Results of the optimal training pathway are marked in bold and highlighted with \colorbox{gray!30}{\phantom{zz}}.}
    \label{tab:compare_transfer}
    \aboverulesep=0.1ex
	\belowrulesep=0.15ex
    \setlength{\tabcolsep}{2pt}
    \scalebox{0.81}{
    \begin{tabular}{lccc}
        \toprule
        Model & setting & Acc\big($\mathcal{M}(\widetilde{\mathcal{A}})$\big) & Acc\big($\mathcal{M}(\widetilde{\mathcal{B}})$\big)\\
        \midrule
        \multirow{5}{*}{ResNet18} & 1 & 96.80 $\pm$ 0.07 & 81.81 $\pm$ 0.53\\
        & \cellcolor{gray!30}\textbf{2} & \cellcolor{gray!30}\textbf{96.38 $\pm$ 0.32} & \cellcolor{gray!30}\textbf{92.32 $\pm$ 0.85}\\
        & 3 & 94.25 $\pm$ 0.83 & 91.49 $\pm$ 0.44\\
        & 4 & 88.02 $\pm$ 1.82 & 90.15 $\pm$ 0.82\\
        & 5 & 96.11 $\pm$ 0.66 & 86.04 $\pm$ 0.79\\
        \midrule
        \multirow{5}{*}{ResNet34} & 1 & 96.58 $\pm$ 0.35 & 79.85 $\pm$ 1.18\\
        & \cellcolor{gray!30}\textbf{2} & \cellcolor{gray!30}\textbf{96.51 $\pm$ 0.40} & \cellcolor{gray!30}\textbf{92.80 $\pm$ 0.23}\\
        & 3 & 94.65 $\pm$ 1.07 & 91.67 $\pm$ 0.62\\
        & 4 & 90.94 $\pm$ 1.15 & 91.16 $\pm$ 0.43\\
        & 5 & 96.81 $\pm$ 0.06 & 88.73 $\pm$ 0.58\\
        \midrule
        \multirow{5}{*}{Inception-v3} & 1 & 96.14 $\pm$ 0.93 & 83.33 $\pm$ 1.27\\
        & \cellcolor{gray!30}\textbf{2} & \cellcolor{gray!30}\textbf{96.77 $\pm$ 0.40} & \cellcolor{gray!30}\textbf{93.11 $\pm$ 0.48}\\
        & 3 & 95.96 $\pm$ 0.68 & 91.88 $\pm$ 0.38\\
        & 4 & 88.69 $\pm$ 0.90 & 90.90 $\pm$ 0.69\\
        & 5 & 96.97 $\pm$ 0.27 & 89.02 $\pm$ 1.13\\
        \bottomrule
    \end{tabular}}
\end{minipage}
\hspace{1mm}
\begin{minipage}[b]{0.63\linewidth}
\centering
\caption{Detailed summary of peak prediction performance yielded by three fine-grained visual classifiers under five training pathways that defined in Eq.~(\ref{eq:setups}). Accuracy is measured in \%. Peak performance under each training pathway is marked in bold and highlighted with \colorbox{gray!30}{\phantom{zz}}. The visual explanations of ResNet34 models with the peak performances are provided in Fig.~\ref{fig:path_compare} as an example.}
	\label{tab:performance_of_5_settings}
	\aboverulesep=0.1ex
	\belowrulesep=0.15ex
    \setlength{\tabcolsep}{2pt}
    \scalebox{0.8}{
	\begin{tabular}{lc|cc|cc}

		\toprule
		Model & setting  &  Acc\big($\mathcal{M}^{\widehat{\mathcal{A}}}(\widetilde{\mathcal{A}}_{\text{F1}}, \widetilde{\mathcal{A}}_{\text{F2}})$\big)  &  Acc\big($\mathcal{M}^{\widehat{\mathcal{A}}}(\widetilde{\mathcal{A}})$\big)
		&
		Acc\big($\mathcal{M}^{\widehat{\mathcal{A}}}(\widetilde{\mathcal{B}}_{\text{F1}}, \widetilde{\mathcal{B}}_{\text{F2}})$\big)  & Acc\big($\mathcal{M}^{\widehat{\mathcal{A}}}(\widetilde{\mathcal{B}})$\big)\\
		\midrule
        \multirow{5}{*}{ResNet18} & 1 & (94.60, 98.78) &	96.69 & (78.64, 86.21)	& 82.42\\
        & \cellcolor{gray!30}\textbf{2} & \cellcolor{gray!30}(\textbf{94.82}, \textbf{98.17}) &	\cellcolor{gray!30}\textbf{96.50} & \cellcolor{gray!30}(\textbf{90.91}, \textbf{96.35}) & \cellcolor{gray!30}\textbf{93.63}\\

        & 3 & (91.79, 99.39) &	95.59 & (83.64, 98.84) & 91.24\\

        & 4 & (91.14, 88.69) &	89.91 & (84.55, 98.34) & 91.44\\

        & 5 & (94.74, 97.55) & 96.14 & (79.55, 92.86) & 86.20\\

        \midrule
		\multirow{5}{*}{ResNet34} & 1 & (95.03, 98.17) & 96.60 & (78.18, 85.55) & 81.87 \\

        & \cellcolor{gray!30}\textbf{2} & \cellcolor{gray!30}(\textbf{95.46}, \textbf{98.17}) & \cellcolor{gray!30}\textbf{96.81} & \cellcolor{gray!30}(\textbf{87.27}, \textbf{98.84}) & \cellcolor{gray!30}\textbf{93.06}\\

        & 3 & (94.53, 96.64) & 95.59 & (84.09, 99.17) & 91.63\\

        & 4 & (92.30, 92.97) & 92.63 & (83.64, 98.67) & 91.16\\

        & 5 & (94.74, 98.78) & 96.76 & (80.45,98.34) & 89.40\\
        \midrule

        \multirow{5}{*}{Inception-v3} & 1 & (94.89, 99.69) & 97.29 & (80.00, 89.37) & 84.69\\

        & \cellcolor{gray!30}\textbf{2} & \cellcolor{gray!30}(\textbf{95.54}, \textbf{98.17}) & \cellcolor{gray!30}\textbf{96.86} & \cellcolor{gray!30}(\textbf{90.91}, \textbf{95.85}) & \cellcolor{gray!30}\textbf{93.38}\\

        & 3 & (95.25, 96.94) & 96.09 & (86.36, 98.50) & 92.43\\

        & 4 & (87.69, 92.0) & 89.87 & (84.55, 98.34) & 91.44\\

        & 5 & (94.96, 99.08) & 97.02 & (84.09,	96.51) & 90.30\\
		
		\bottomrule
	\end{tabular}
 }
\end{minipage}
\end{table*}

Noteworthy, the training pathways 1 and 2 are similar to 4 and 5\textcolor{red}{,} where 1 and 3 are the starting point of the proposed DCF\textcolor{red}{,} each of which was trained on with \texttt{zero} padding scheme. The third one is different among the five training pathways as it was trained on a combination of both training sets with \texttt{reflection} padding scheme applied directly.


Apart from ResNet34, which was trained as the backbone model in training setting 1, we also employed ResNet18 and Inception-v3 in such a setting. The detailed performance is reported in Table~\ref{tab:compare_transfer}, where the model with peak prediction performances is highlighted and used for the fine-tuning process required by training setting 2.

\begin{figure}[ht!]
  \centering
  \includegraphics[width=0.98\linewidth]{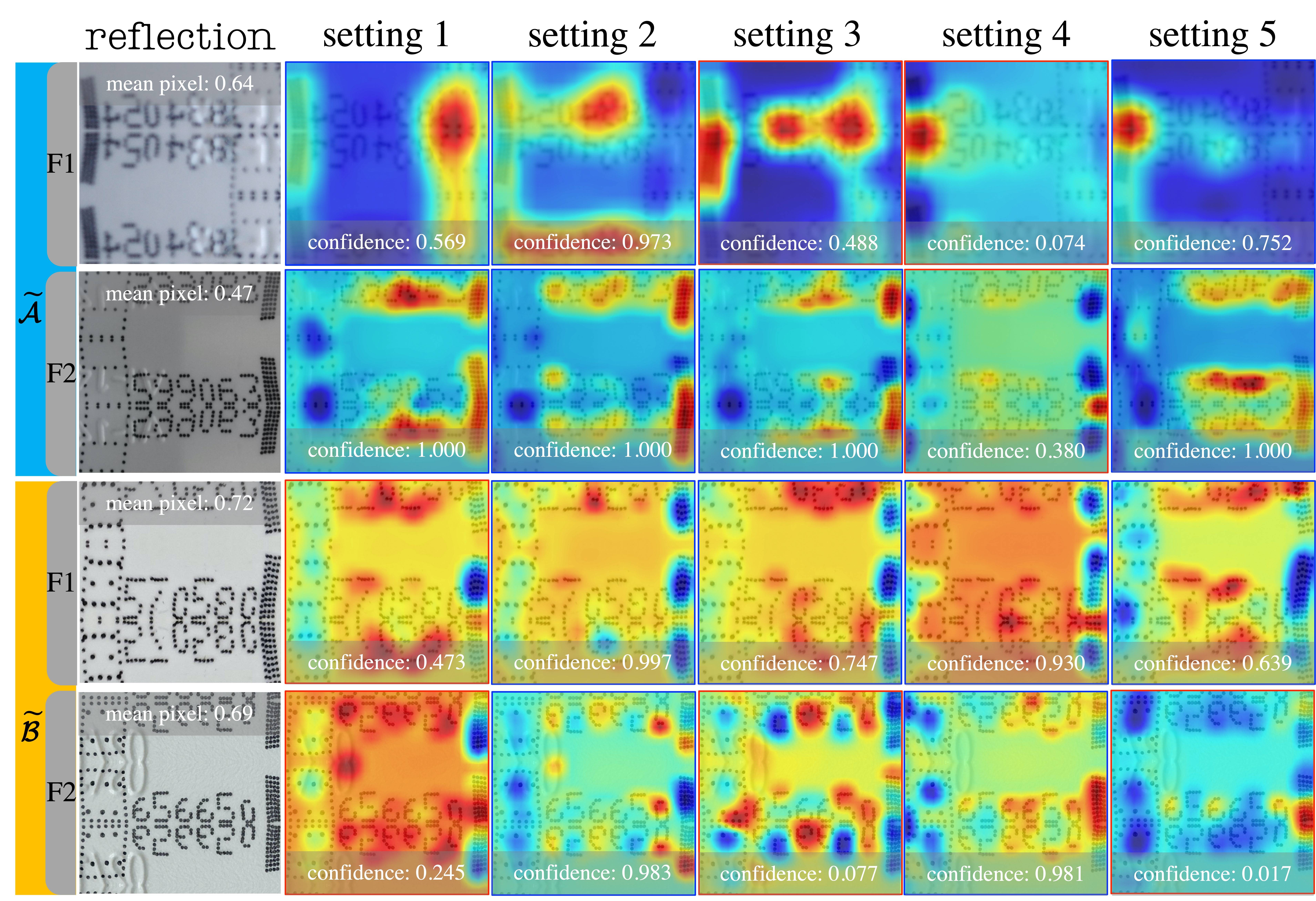}
  \caption{Visual explanations supported by five ResNet34 models trained using various training pathways as summarised in Table~\ref{tab:performance_of_5_settings}. The adopted colour codes are consistent with Fig.~\ref{fig:padding_exp}. The second training pathway (\emph{i.e.}~setting 2) has demonstrated its compatibility with blurry and clean images (first two rows) and illustrated its discriminability in coping with finer patterns,~\emph{e.g.} printed dents (last row) in the testing sets of $\mathcal{A}$ and $\mathcal{B}$ (\emph{i.e.}~$\widetilde{\mathcal{A}}$ and $\widetilde{\mathcal{B}}$).}
   \label{fig:path_compare}
\end{figure}

Unlike training settings 2 and 4, we train a model using both training sets from scratch with the qualified padding scheme (\emph{i.e.}~\texttt{reflection} padding). In the meantime, as the peak performance presented in Table~\ref{tab:performance_of_5_settings} and partially explained in Fig.~\ref{fig:path_compare}, the prediction performance resulting from training setting 2 is much better than that of setting 4, emphasising that the order of training datasets matters.

Overall, the outperformance of training setting 2 not only reveals that different padding schemes could compensate with each other but also addresses the possible overfitting of training data~\cite{guo2020adafilter} and catastrophic forgetting of model~\cite{kong2023overcoming}.

\section{Conclusion}
\label{sec:con}
In this paper, we proposed an optimal training pathway selection framework DCF for robust and explainable fine-grained visual classification in the context where two temporal continued datasets are provided, and a model has been pre-trained on the dataset released earlier. The DCF consists of two adjacent steps,~\emph{i.e.} determining the optimal padding scheme and selecting the optimal training pathway, each equipped with a quantitative and visual explanation in a \emph{putting-through} fashion. The experimental results first indicate that the best-suited padding scheme determination is a feasible starting point for deploying the proposed DCF, as it reveals great potential in enhancing the eventual classification performance. The efficiency and efficacy of DCF are then confirmed by its second step of figuring out the optimal training pathway, where the classification accuracy over two datasets is maximally enhanced and leveraged. Experimental results also inspire further investigations on image-wise quality separation~\cite{img_quality_2024} of the optimal learning pathway selection for boosted image categorisation with mitigated data bias~\cite{qraitem2023bias} and reduced computational cost. 

\noindent \textbf{Acknowledgment.} $\quad$ This work was supported by Procter \& Gamble (P\&G) United Kingdom Technical Centres Ltd.

\onecolumn

\begin{center}

\section*{\begin{LARGE}Supplementary material for ``Robust and Explainable Fine-Grained Visual Classification with Transfer Learning: A Dual-Carriageway Framework''\end{LARGE}}
\end{center}

\vspace{6em}
\setcounter{page}{1}
\section*{A. Optimal Padding Scheme Determination}
\label{sec:opsd}

\begin{table*}[!ht]
    \centering
\caption{Detailed prediction performance (in \%) comparisons on the testing sets of datasets $\mathcal{A}$ and $\mathcal{B}$, \emph{i.e.}~$\widetilde{\mathcal{A}}$ and $\widetilde{\mathcal{B}}$ regarding the labels `F1' and `F2', using the ResNet34~\cite{he2015deep} (\emph{i.e.}~$\mathcal{M}$) trained on the training set of dataset $\mathcal{A}$ with \texttt{zero} padding (\emph{i.e.} $\widehat{\mathcal{A}}$) that was processed by each of the six padding schemes. Results yielded by the optimal padding scheme are marked in bold and highlighted with \colorbox{gray!30}{\phantom{zz}}.}
	\label{tab:full_details_pad}
	\aboverulesep=0.1ex
	\belowrulesep=0.15ex
    \setlength{\tabcolsep}{2pt}
    \scalebox{0.96}{
	\begin{tabular}{lc|ccc|ccc}

		\toprule
		Padding Scheme & run  &  Acc\big($\mathcal{M}^{\widehat{\mathcal{A}}}(\widetilde{\mathcal{A}}_{\text{F1}}, \widetilde{\mathcal{A}}_{\text{F2}})$\big)  &  Acc\big($\mathcal{M}^{\widehat{\mathcal{A}}}(\widetilde{\mathcal{A}})$\big) & mean $\pm$ std.
		&
		Acc\big($\mathcal{M}^{\widehat{\mathcal{A}}}(\widetilde{\mathcal{B}}_{\text{F1}}, \widetilde{\mathcal{B}}_{\text{F2}})$\big)  & Acc\big($\mathcal{M}^{\widehat{\mathcal{A}}}(\widetilde{\mathcal{B}})$\big) & mean $\pm$ std.\\
		\midrule
  
        \multirow{5}{*}{\texttt{zero}} & 1 & (93.16, 96.02) & 94.59 & \multirow{5}{*}{94.05 $\pm$ 2.03} & (65.91, 80.73)	& 73.32 & \multirow{5}{*}{71.79 $\pm$ 3.15}\\
        & 2 & (92.58, 97.55) &	95.06 & & (71.36, 76.74)	& 74.05 &\\
        & 3 & (95.25, 85.63) &	90.44 & & (82.73, 50.17)	& 66.45 &\\
        & 4 & (94.17, 95.72) &	94.94 & & (77.27, 65.61)	& 71.44 &\\
        & \cellcolor{gray!30}\textbf{5} & \cellcolor{gray!30}(\textbf{95.03}, \textbf{95.41}) &	\cellcolor{gray!30}\textbf{95.22} & & \cellcolor{gray!30}(\textbf{74.09}, \textbf{73.26})	& \cellcolor{gray!30}\textbf{73.68} &\\

        \midrule
		\multirow{5}{*}{\texttt{RGB-mean}} & 1 & (94.17, 96.02) &	95.09 & \multirow{5}{*}{95.07 $\pm$ 0.53} & (73.18, 73.59)	& 73.39 & \multirow{5}{*}{74.59 $\pm$ 0.86}\\
        & 2 & (94.46, 93.88) &	94.17 & & (75.00, 73.75)	& 74.38 &\\
        & \cellcolor{gray!30}\textbf{3} & \cellcolor{gray!30}(\textbf{94.10}, \textbf{96.64}) &	\cellcolor{gray!30}\textbf{95.37} & & \cellcolor{gray!30}(\textbf{77.73}, \textbf{73.59})	& \cellcolor{gray!30}\textbf{75.66} &\\
        & 4 & (94.31, 96.02) &	95.16 & & (76.82, 71.93)	& 74.38 &\\
        & 5 & (94.46, 96.64) &	95.55 & & (74.55, 75.75)	& 75.15 &\\
        \midrule

        \multirow{5}{*}{\texttt{LAB-mean}} & 1 & (95.10, 96.02) &	95.56 & \multirow{5}{*}{95.59 $\pm$ 0.48} & (75.00, 68.44)	& 71.72 & \multirow{5}{*}{72.61 $\pm$ 1.21}\\
        & 2 & (94.74, 96.02) &	95.38 && (70.91, 73.92)	& 72.41&\\
        & 3 & (94.67, 97.86) &	96.27 && (75.00, 70.10)	& 72.55&\\
        & 4 & (93.38, 98.17) &	95.78 && (75.00, 68.44)	& 71.72&\\
        & \cellcolor{gray!30}\textbf{5} & \cellcolor{gray!30}(\textbf{94.24}, \textbf{95.72}) &	\cellcolor{gray!30}\textbf{94.98} && \cellcolor{gray!30}(\textbf{72.73}, \textbf{76.58})	& \cellcolor{gray!30}\textbf{74.66}&\\
		\midrule
  
        \multirow{5}{*}{\texttt{white}} & 1 & (93.09, 96.94) &	95.02 & \multirow{5}{*}{94.89 $\pm$ 0.20} & (75.00, 77.41)	& 76.20 & \multirow{5}{*}{75.78 $\pm$ 1.41}\\
        & \cellcolor{gray!30}\textbf{2} & \cellcolor{gray!30}(\textbf{94.74}, \textbf{95.41}) &	\cellcolor{gray!30}\textbf{95.07} && \cellcolor{gray!30}(\textbf{75.00}, \textbf{79.73})	& \cellcolor{gray!30}\textbf{77.37}&\\
        & 3 & (94.82, 95.11) &	94.97 && (73.18, 76.58)	& 74.88&\\
        & 4 & (93.81, 95.72) &	94.77 && (75.00, 78.24)	& 76.62&\\
        & 5 & (94.10, 95.11) &	94.60 && (70.45, 77.24)	& 73.84&\\

        \midrule
		\multirow{5}{*}{\texttt{grey}} & 1 & (94.67, 96.64) &	95.66 & \multirow{5}{*}{95.10 $\pm$ 0.41} & (79.55, 74.42)	& 76.98 & \multirow{5}{*}{77.10 $\pm$ 1.06}\\
        & 2 & (93.74, 95.41) &	94.57 && (75.91, 78.90)	& 77.41&\\
        & \cellcolor{gray!30}\textbf{3} & \cellcolor{gray!30}(\textbf{93.81}, \textbf{96.02}) &	\cellcolor{gray!30}\textbf{94.91} && \cellcolor{gray!30}(\textbf{74.55}, \textbf{81.89})	& \cellcolor{gray!30}\textbf{78.22}&\\
        & 4 & (94.74, 95.41) &	95.07 && (75.00, 75.75)	& 75.38&\\
        & 5 & (94.60, 96.02) &	95.31 && (74.09, 80.90)	& 77.50&\\
        \midrule

        \multirow{5}{*}{\texttt{reflection}} & 1 & (95.10, 99.08) &	97.09 & \multirow{5}{*}{\textbf{96.58 $\pm$ 0.35}} & (75.91, 83.72)	& 79.81 & \multirow{5}{*}{\textbf{79.85 $\pm$ 1.18}}\\
        & 2 & (95.18, 98.17) &	96.68 && (77.73, 80.23)	& 78.98&\\
        & \cellcolor{gray!30}\textbf{3} & \cellcolor{gray!30}(\textbf{95.03}, \textbf{98.17}) &	\cellcolor{gray!30}\textbf{96.60} & \cellcolor{gray!30}\textbf{96.58 $\pm$ 0.35} & \cellcolor{gray!30}(\textbf{78.18}, \textbf{85.55})	& \cellcolor{gray!30}\textbf{81.87}& \cellcolor{gray!30}\textbf{79.85 $\pm$ 1.18}\\
        & 4 & (94.82, 97.55) &	96.19 && (78.18, 79.90)	& 79.04&\\
        & 5 & (95.10, 97.55) &	96.32 && (77.73, 81.40)	& 79.56&\\
		\bottomrule
	\end{tabular}
 }
\end{table*}

\vspace{6em}

\begin{figure*}[!ht]
        \centering
        \includegraphics[width=\linewidth]{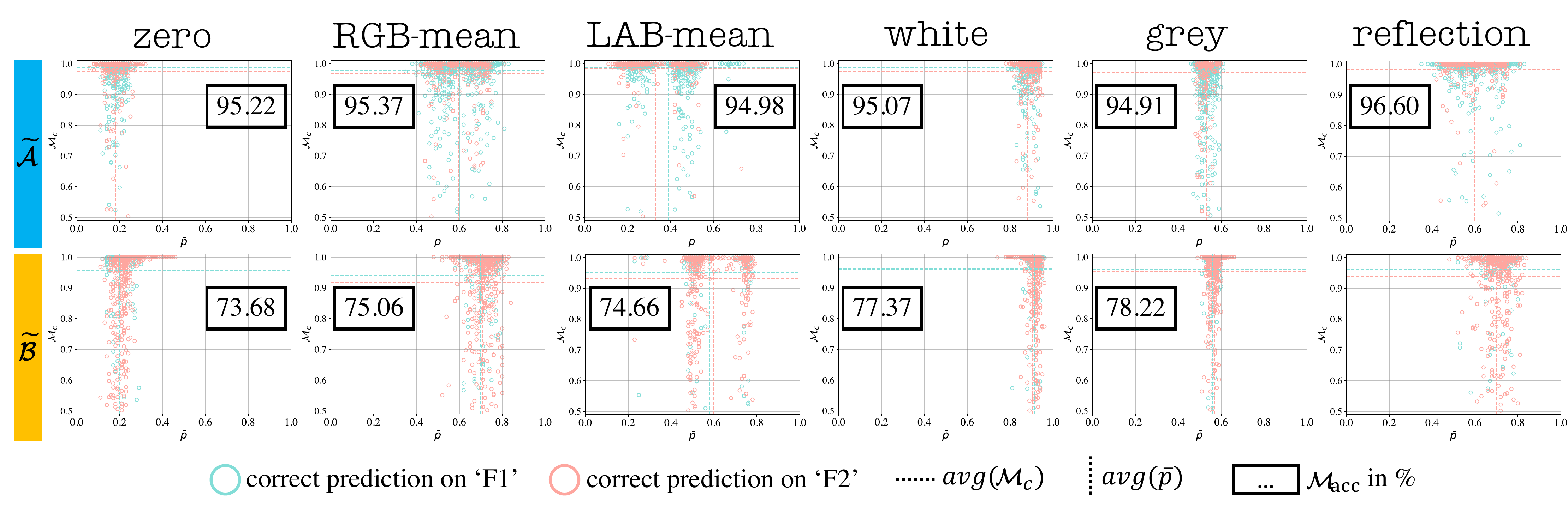}
        \captionof{figure}{Quantitative explanation of the relationship between input pixel value and confidence of correct predictions given by the ResNet34 under varying padding schemes presented in Table~\ref{tab:full_details_pad}. $avg(\cdot)$ calculates the average for the element regarding the total number of correctly predicted samples of the subset (grouped by label) within the testing set, $\bar{p}$ (normalised to the range of $[0,1]$) represents mean pixel value, $\mathcal{M}_{c}$ denotes model confidence and $\mathcal{M}_{\text{acc}}$ corresponds to model prediction accuracy (in \%). Best viewed in colour and zoomed mode.}
        \label{fig:quanti_analysis_visual}
 \end{figure*}

\section*{B. Optimal Training Pathway Selection}

\begin{table*}[!ht]
    \centering
\caption{Detailed prediction performance (in \%) comparisons on the testing sets of datasets $\mathcal{A}$ and $\mathcal{B}$, \emph{i.e.}~$\widetilde{\mathcal{A}}$ and $\widetilde{\mathcal{B}}$, using the ResNet18~\cite{he2015deep} trained on the training set of dataset $\mathcal{A}$ (\emph{i.e.} $\widehat{\mathcal{A}}$) with \texttt{reflection} padding (as summarised in Table~\ref{tab:full_details_pad}) under five different training pathways in the proposed DCF. Results yielded by the optimal padding scheme are marked in bold and highlighted with \colorbox{gray!30}{\phantom{zz}}.}
	\label{tab:full_details_pathway_resnet18}
	\aboverulesep=0.1ex
	\belowrulesep=0.15ex
    \setlength{\tabcolsep}{2pt}
    \scalebox{0.98}{
	\begin{tabular}{lcc|ccc|ccc}

		\toprule
		Model & setting & run  &  Acc\big($\mathcal{M}^{\widehat{\mathcal{A}}}(\widetilde{\mathcal{A}}_{\text{F1}}, \widetilde{\mathcal{A}}_{\text{F2}})$\big)  &  Acc\big($\mathcal{M}^{\widehat{\mathcal{A}}}(\widetilde{\mathcal{A}})$\big) & mean $\pm$ std.
		&
		Acc\big($\mathcal{M}^{\widehat{\mathcal{A}}}(\widetilde{\mathcal{B}}_{\text{F1}}, \widetilde{\mathcal{B}}_{\text{F2}})$\big)  & Acc\big($\mathcal{M}^{\widehat{\mathcal{A}}}(\widetilde{\mathcal{B}})$\big) & mean $\pm$ std.\\
		\midrule
  
        \multirow{25}{*}{\rotatebox{90}{ResNet18}} & \multirow{5}{*}{1} & \cellcolor{gray!30}\textbf{1} & \cellcolor{gray!30}(\textbf{94.60}, \textbf{98.78}) & \cellcolor{gray!30}\textbf{96.69} & \multirow{5}{*}{96.80 $\pm$ 0.07} & \cellcolor{gray!30}(\textbf{78.64}, \textbf{86.21})	& \cellcolor{gray!30}\textbf{82.42} & \multirow{5}{*}{81.81 $\pm$ 0.53}\\
        & & 2 & (94.53, 99.08) & 96.81 &  & (81.36, 83.06)	& 82.21 & \\
        & & 3 & (94.67, 99.08) & 96.88 &  & (76.36, 86.88)	& 81.62 & \\
        & & 4 & (94.53, 99.08) & 96.81 &  & (79.55, 82.56)	& 81.06 & \\
        & & 5 & (93.95, 99.69) & 96.82 &  & (77.27, 86.21)	& 81.74 & \\

        \cline{2-9}
		& & 1 & (95.61, 96.33) & 95.97 & & (84.09, 99.34)	& 91.72 & \\
        & & 2 & (95.03, 97.25) & 96.14 &  & (87.27, 98.17)	& 92.72 & \\
        & \cellcolor{gray!30}\textbf{2} & 3 & (94.74, 98.47) & 96.60 & \cellcolor{gray!30}\textbf{96.38 $\pm$ 0.32} & (84.55, 98.84)	& 91.69 & \cellcolor{gray!30}\textbf{92.32 $\pm$ 0.85}\\
        & & 4 & (95.25, 98.17) & 96.71 &  & (85.00, 98.67)	& 91.84 & \\
        & & \cellcolor{gray!30}\textbf{5} & \cellcolor{gray!30}(\textbf{94.82}, \textbf{98.17}) & \cellcolor{gray!30}\textbf{96.50} &  & \cellcolor{gray!30}(\textbf{90.91}, \textbf{96.35})	& \cellcolor{gray!30}\textbf{93.63} & \\
        \cline{2-9}

        & \multirow{5}{*}{3} & 1 & (91.79, 96.64) & 94.22 & \multirow{5}{*}{94.25 $\pm$ 0.83} & (82.73, 99.00)	& 90.87 & \multirow{5}{*}{91.49 $\pm$ 0.44}\\
        & & 2 & (91.79, 96.02) & 93.91 &  & (84.09, 99.17)	& 91.63 & \\
        & & 3 & (92.01, 96.33) & 94.17 &  & (84.55, 98.84)	& 91.69 & \\
        & & \cellcolor{gray!30}\textbf{4} & \cellcolor{gray!30}(\textbf{91.79}, \textbf{99.39}) & \cellcolor{gray!30}\textbf{95.59} &  & \cellcolor{gray!30}(\textbf{83.64}, \textbf{98.84})	& \cellcolor{gray!30}\textbf{91.24} & \\
        & & 5 & (91.29, 95.41) & 93.35 &  & (85.00, 99.00)	& 92.00 & \\
		\cline{2-9}
  
        & \multirow{5}{*}{4} & 1 & (90.78, 86.54) & 88.66 & \multirow{5}{*}{88.02 $\pm$ 1.82} & (80.91, 98.50)	& 89.70 & \multirow{5}{*}{90.15 $\pm$ 0.82}\\
        & & \cellcolor{gray!30}\textbf{2} & \cellcolor{gray!30}(\textbf{91.14}, \textbf{88.69}) & \cellcolor{gray!30}\textbf{89.91} &  & \cellcolor{gray!30}(\textbf{84.55}, \textbf{98.34})	& \cellcolor{gray!30}\textbf{91.44} & \\
        & & 3 & (92.51,	85.02) & 88.77 &  & (80.91, 98.34)	& 89.62 & \\
        & & 4 & (90.06, 80.12) & 85.09 &  & (82.27, 98.67)	& 90.47 & \\
        & & 5 & (90.06, 85.32) & 87.69 &  & (81.82, 97.18)	& 89.50 & \\

        \cline{2-9}
		& \multirow{5}{*}{5} & 1 & (95.10, 98.47) & 96.78 & \multirow{5}{*}{96.11 $\pm$ 0.66} & (79.09, 91.03)	& 85.06 & \multirow{5}{*}{86.04 $\pm$ 0.79}\\
        & & 2 & (94.17, 97.55) & 95.86 &  & (81.36, 91.36)	& 86.36 & \\
        & & 3 & (95.10, 98.17) & 96.63 &  & (77.27, 93.69)	& 85.48 & \\
        & & 4 & (92.73, 97.55) & 95.14 &  & (78.64, 95.51)	& 87.08 & \\
        & & \cellcolor{gray!30}\textbf{5} & \cellcolor{gray!30}(\textbf{94.74}, \textbf{97.55}) & \cellcolor{gray!30}\textbf{96.14} &  & \cellcolor{gray!30}(\textbf{79.55}, \textbf{92.86})	& \cellcolor{gray!30}\textbf{86.20} & \\
		\bottomrule
	\end{tabular}
 }
\end{table*}

\begin{table*}
    \centering
\caption{Detailed prediction performance (in \%) comparisons on the testing sets of datasets $\mathcal{A}$ and $\mathcal{B}$ using the ResNet34~\cite{he2015deep} and Inception-v3~\cite{szegedy2016rethinking} trained on the training set of dataset $\mathcal{A}$ with \texttt{reflection} padding under five different training pathways. Results yielded by the optimal padding scheme are marked in bold and highlighted with \colorbox{gray!30}{\phantom{zz}}.}
	\label{tab:full_details_pathway_resnet34}
	\aboverulesep=0.1ex
	\belowrulesep=0.15ex
    \setlength{\tabcolsep}{2pt}
    \scalebox{0.98}{
	\begin{tabular}{lcc|ccc|ccc}

		\toprule
		Model & setting & run  &  Acc\big($\mathcal{M}^{\widehat{\mathcal{A}}}(\widetilde{\mathcal{A}}_{\text{F1}}, \widetilde{\mathcal{A}}_{\text{F2}})$\big)  &  Acc\big($\mathcal{M}^{\widehat{\mathcal{A}}}(\widetilde{\mathcal{A}})$\big) & mean $\pm$ std.
		&
		Acc\big($\mathcal{M}^{\widehat{\mathcal{A}}}(\widetilde{\mathcal{B}}_{\text{F1}}, \widetilde{\mathcal{B}}_{\text{F2}})$\big)  & Acc\big($\mathcal{M}^{\widehat{\mathcal{A}}}(\widetilde{\mathcal{B}})$\big) & mean $\pm$ std.\\
		\midrule
  
        \multirow{25}{*}{\rotatebox{90}{ResNet34}} & \multirow{5}{*}{1} & 1 & (95.10, 99.08) & 97.09 &  & (75.91, 83.72)	& 79.81 & \\
        & & 2 & (95.18, 98.17) & 96.68 &  & (77.73, 80.23)	& 78.98 & \\
        & & \cellcolor{gray!30}\textbf{3} & \cellcolor{gray!30}(\textbf{95.03}, \textbf{98.17}) & \cellcolor{gray!30}\textbf{96.60} &  96.58 $\pm$ 0.35 & \cellcolor{gray!30}(\textbf{78.18}, \textbf{85.55})	& \cellcolor{gray!30}\textbf{81.87} & 79.85 $\pm$ 1.18\\
        & & 4 & (94.82, 97.55) & 96.19 &  & (78.18, 79.90)	& 79.04 & \\
        & & 5 & (95.10, 97.55) & 96.32 &  & (77.73, 81.40)	& 79.56 & \\

        \cline{2-9}
		& & 1 & (95.25, 98.47) & 96.86 &  & (85.91, 99.17)	& 92.54 & \\
        & & 2 & (95.46,	97.25) & 96.35 &  & (90.00, 95.85)	& 92.92 & \\
        & \cellcolor{gray!30}\textbf{2} & 3 & (95.46,	97.86) & 96.66 & \cellcolor{gray!30}\textbf{96.51 $\pm$ 0.40} & (87.27, 97.84)	& 92.56 & \cellcolor{gray!30}\textbf{92.80 $\pm$ 0.23}\\
        & & 4 & (95.46,	96.33) & 95.89 &  & (88.64, 97.18)	& 92.91 & \\
        & & \cellcolor{gray!30}\textbf{5} & \cellcolor{gray!30}(\textbf{95.46},	\textbf{98.17}) & \cellcolor{gray!30}\textbf{96.81} &  & \cellcolor{gray!30}(\textbf{87.27},	\textbf{98.84})	& \cellcolor{gray!30}\textbf{93.06} & \\
        \cline{2-9}

        & \multirow{5}{*}{3} & 1 & (94.02, 96.64) & 95.33 & \multirow{5}{*}{94.65 $\pm$ 1.07} & (85.00, 98.67)	& 91.84 & \multirow{5}{*}{91.67 $\pm$ 0.62}\\
        & & \cellcolor{gray!30}\textbf{2} & \cellcolor{gray!30}(\textbf{94.53},	\textbf{96.64}) & \cellcolor{gray!30}\textbf{95.59} &  & \cellcolor{gray!30}(\textbf{84.09}, \textbf{99.17})	& \cellcolor{gray!30}\textbf{91.63} & \\
        & & 3 & (93.38,	93.88) & 93.63 &  & (83.64, 98.17)	& 90.91 & \\
        & & 4 & (92.80,	97.86) & 95.33 &  & (84.09, 98.67)	& 91.38 & \\
        & & 5 & (94.96,	91.74) & 93.35 &  & (86.36, 98.84)	& 92.60 & \\
		\cline{2-9}
  
        & \multirow{5}{*}{4} & 1 & (87.90,	94.19) & 91.05 & \multirow{5}{*}{90.94 $\pm$ 1.15}  & (84.09,	98.17)	& 91.13 & \multirow{5}{*}{91.16 $\pm$ 0.43}\\
        & & 2 & (89.27,	93.27) & 91.27 &  & (82.27,	98.67)	& 90.47 & \\
        & & \cellcolor{gray!30}\textbf{3} & \cellcolor{gray!30}(\textbf{92.30},	\textbf{92.97}) & \cellcolor{gray!30}\textbf{92.63} &  & \cellcolor{gray!30}(\textbf{83.64},	\textbf{98.67})	& \cellcolor{gray!30}\textbf{91.16} & \\
        & & 4 & (89.63,	90.52) & 90.07 &  & (85.45,	97.67)	& 91.56 & \\
        & & 5 & (90.42,	88.99) & 89.70 &  & (85.45,	97.51)	& 91.48 & \\

        \cline{2-9}
		& \multirow{5}{*}{5} & 1 & (94.67,	99.08) & 96.88 & \multirow{5}{*}{96.81 $\pm$ 0.06}  & (80.00,	96.18)	& 88.09 & \multirow{5}{*}{88.73 $\pm$ 0.58}\\
        & & 2 & (94.96,	98.78) & 96.87 &  & (80.00,	97.18)	& 88.59 & \\
        & & 3 & (95.03,	98.47) & 96.75 &  & (81.36,	97.18)	& 89.27 & \\
        & & 4 & (95.10,	98.47) & 96.78 &  & (79.09,	97.51)	& 88.30 & \\
        & & \cellcolor{gray!30}\textbf{5} & \cellcolor{gray!30}(\textbf{94.74},	\textbf{98.78}) & \cellcolor{gray!30}\textbf{96.76} &  & \cellcolor{gray!30}(\textbf{80.45},	\textbf{98.34})	& \cellcolor{gray!30}\textbf{89.40} & \\
		\bottomrule

        \multirow{25}{*}{\rotatebox{90}{Inception-v3}} & \multirow{5}{*}{1} & 1 & (94.82,	97.86) & 96.34 & \multirow{5}{*}{96.14 $\pm$ 0.93} & (77.27,	87.38)	& 82.32 & \multirow{5}{*}{83.33 $\pm$ 1.27}\\
        & & \cellcolor{gray!30}\textbf{2} & \cellcolor{gray!30}(\textbf{94.89},	\textbf{99.69}) & \cellcolor{gray!30}\textbf{97.29} &  & \cellcolor{gray!30}(\textbf{80.00},	\textbf{89.37})	& \cellcolor{gray!30}\textbf{84.69} & \\
        & & 3 & (94.38,	97.55) & 95.97 &  & (77.73,	89.87)	& 83.80 & \\
        & & 4 & (93.45,	96.02) & 94.73 &  & (79.09,	89.20)	& 84.15 & \\
        & & 5 & (94.89,	97.86) & 96.38 &  & (76.36,	87.04)	& 81.70 & \\

        \cline{2-9}
		&  & 1 & (95.82,	97.55) & 96.69 &   & (86.82,	98.67)	& 92.75 & \\
        & & 2 & (95.03,	97.86) & 96.44 &  & (88.64,	97.67)	& 93.16 & \\
        & \cellcolor{gray!30}\textbf{2} & \cellcolor{gray!30}\textbf{3} & \cellcolor{gray!30}(\textbf{95.54},	\textbf{98.17}) & \cellcolor{gray!30}\textbf{96.86} & \cellcolor{gray!30}\textbf{96.77 $\pm$ 0.40} & \cellcolor{gray!30}(\textbf{90.91}, \textbf{95.85})	& \cellcolor{gray!30}\textbf{93.38} & \cellcolor{gray!30}\textbf{93.11 $\pm$ 0.48}\\
        & & 4 & (95.32,	97.55) & 96.44 &  & (88.64,	98.84)	& 93.74 & \\
        & & 5 & (95.75,	99.08) & 97.41 &  & (87.73,	97.34)	& 92.53 & \\
        \cline{2-9}

        & \multirow{5}{*}{3} & 1 & (95.39,	94.80) & 95.09 &  \multirow{5}{*}{95.96 $\pm$ 0.68}& (84.55,	99.00)	& 91.78 & \multirow{5}{*}{91.88 $\pm$ 0.38}\\
        & & \cellcolor{gray!30}\textbf{2} & \cellcolor{gray!30}(\textbf{95.25},	\textbf{96.94}) & \cellcolor{gray!30}\textbf{96.09} &  & \cellcolor{gray!30}(\textbf{86.36},	\textbf{98.50})	& \cellcolor{gray!30}\textbf{92.43} & \\
        & & 3 & (94.10,	98.78) & 96.44 &  & (85.45,	98.50)	& 91.97 & \\
        & & 4 & (94.89,	96.02) & 95.45 &  & (84.55,	99.17)	& 91.86 & \\
        & & 5 & (94.38,	99.08) & 96.73 &  & (84.55,	98.17)	& 91.36 & \\
		\cline{2-9}
  
        & \multirow{5}{*}{4} & 1 & (93.81,	83.49) & 88.65 & \multirow{5}{*}{88.69 $\pm$ 0.90} & (84.09,	98.17)	& 91.13 & \multirow{5}{*}{90.90 $\pm$ 0.69}\\
        & & 2 & (85.17,	89.60) & 87.38 &  & (85.00,	97.18)	& 91.09 & \\
        & & 3 & (92.01,	85.02) & 88.52 &  & (83.64, 98.67)	& 91.16 & \\
        & & 4 & (90.64,	87.46) & 89.05 &  & (81.36,	98.01)	& 89.69 & \\
        & & \cellcolor{gray!30}\textbf{5} & \cellcolor{gray!30}(\textbf{87.69},	\textbf{92.05}) & \cellcolor{gray!30}\textbf{89.87} &  & \cellcolor{gray!30}(\textbf{84.55},	\textbf{98.34})	& \cellcolor{gray!30}\textbf{91.44} & \\

        \cline{2-9}
		& \multirow{5}{*}{5} & \cellcolor{gray!30}\textbf{1} & \cellcolor{gray!30}(\textbf{94.96},	\textbf{99.08}) & \cellcolor{gray!30}\textbf{97.02} & \multirow{5}{*}{96.97 $\pm$ 0.27} & \cellcolor{gray!30}(\textbf{84.09}, \textbf{96.51})	& \cellcolor{gray!30}\textbf{90.30} & \multirow{5}{*}{89.02 $\pm$ 1.13}\\
        & & 2 & (94.67,	99.69) & 97.18 &  & (78.64,	96.51)	& 87.58 & \\
        & & 3 & (94.74,	98.47) & 96.60 &  & (82.27,	94.02)	& 88.14 & \\
        & & 4 & (94.82,	99.69) & 97.25 &  & (81.82,	96.84)	& 89.33 & \\
        & & 5 & (95.39,	98.17) & 96.78 &  & (83.64,	95.85)	& 89.75 & \\
		\bottomrule
	\end{tabular}
 }
\end{table*}

\twocolumn
\vspace{10em}
{
    \small
    \bibliographystyle{ieeenat_fullname}
    \bibliography{main}
}

\end{document}